\documentclass[letterpaper, 10 pt, journal, twoside]{IEEEtran}

\usepackage[acronym]{glossaries}
\usepackage{array,multirow,graphics}
\usepackage{graphicx}
\usepackage{times}
\usepackage{amsmath}
\usepackage{amssymb}
\usepackage{amsbsy}
\usepackage[font=footnotesize]{caption}
\usepackage{float}
\usepackage{balance}
\usepackage{url}
\usepackage[ruled,linesnumbered,longend]{algorithm2e}
\usepackage{xcolor}
\usepackage{latexsym}
\usepackage{subcaption}
\usepackage{textcomp}
\usepackage{pdfrender}
\usepackage{tabularx}
\usepackage{booktabs}
\usepackage{mathptmx} 
\usepackage{amsfonts}
\usepackage{makecell}
\usepackage{fancyhdr}
\let\oldmaketitle\maketitle
 
\renewcommand{\maketitle}{%
  \oldmaketitle
  \thispagestyle{fancy}
}

\DeclareMathAlphabet{\mathcal}{OMS}{cmsy}{m}{n}

\def\secref#1{Sec.~\ref{#1}}
\def\figref#1{Fig.~\ref{#1}}
\def\tabref#1{Tab.~\ref{#1}}
\def\eqref#1{Eq.~(\ref{#1})}

\def\ourmethod{AgriColMap}

\newcommand{\norm}[1]{\left\lVert#1\right\rVert}

\title{\ourmethod: Aerial-Ground Collaborative 3D Mapping for Precision Farming}

\newacronym{uav}{UAV}{Unmanned Aerial Vehicle}
\newacronym{ugv}{UGV}{Unmanned Ground Vehicle}
\newacronym{cnn}{CNN}{Convolutional Neural Network}
\newacronym{lidar}{LiDAR}{Light Detection and Ranging}
\newacronym{icp}{ICP}{Iterative Closest Point}
\newacronym{slam}{SLAM}{Simultaneous Localization and Mapping}
\newacronym{ahrs}{AHRS}{Attitude and Heading Reference System}
\newacronym{exg}{ExG}{Excess Green}
\newacronym{fpfh}{FPFH}{Fast Point Feature Histograms}
\newacronym{sift}{SIFT}{Scale-Invariant Feature Transform} 
\newacronym{cpd}{CPD}{coherent point drift}
\newacronym{goicp}{Go-ICP}{Globally Optimal 3D ICP}
\newacronym{gps}{GPS}{Global Positioning System}
\newacronym{dsm}{DSM}{Digital Surface Model} 
\newacronym{ldof}{LDOF}{Large displacement Dense Optical Flow}
\newacronym{cpm}{CPM}{Coarse-to-fine PatchMatch}

\begin{document}


\author{Ciro Potena$^1$, Raghav Khanna$^2$, Juan Nieto$^2$, Roland Siegwart$^2$, Daniele Nardi$^1$, and Alberto Pretto$^1$
\thanks{Manuscript received: September, 10, 2018; Revised November, 07, 2018; Accepted
January, 02, 2019. This paper was recommended for publication by Editor Youngjin Choi upon evaluation of the Associate Editor and Reviewers` comments.}
\thanks{This work was supported by the EC under Grant H2020-ICT-644227-Flourish and by the Swiss State Secretariat for Education, Research and Innovation under contract number 15.0029. $^{1}$Potena, Nardi and Pretto are with the Department of Computer, Control, and Management Engineering ``Antonio Ruberti``, Sapienza University of Rome, Italy. Email: { \{potena, nardi, pretto\}@diag.uniroma1.it}. $^{2}$Khanna, Nieto and Siegwart are with the Autonomous Systems Lab, ETH Zurich, Switzerland.
Email: raghav.khanna@mavt.ethz.ch, jnieto@ethz.ch,r.siegwart@ieee.org}%
\thanks{Digital Object Identifier (DOI): 10.1109/LRA.2019.2894468.}
}

\maketitle

\begin{abstract}

The combination of aerial survey capabilities of \acrlong{uav}s with targeted intervention abilities of agricultural \acrlong{ugv}s can significantly improve the effectiveness of robotic systems applied to precision agriculture. In this context, building and updating a common map of the field is an essential but challenging task. The maps built using robots of different types show differences in size, resolution and scale, the associated geolocation data may be inaccurate and biased, while the repetitiveness of both visual appearance and geometric structures found within agricultural contexts render classical map merging techniques ineffective.\\ 
In this paper we propose \ourmethod, a novel map registration pipeline that leverages a grid-based multimodal environment representation which includes a vegetation index map and a \acrlong{dsm}. We cast the data association problem between maps built from \acrshort{uav}s and \acrshort{ugv}s as a multimodal, large displacement dense optical flow estimation. The dominant, coherent flows, selected using a voting scheme, are used as point-to-point correspondences to infer a preliminary non-rigid alignment between the maps. A final refinement is then performed, by exploiting only meaningful parts of the registered maps.\\
We evaluate our system using real world data for 3 fields with different crop species. The results show that our method outperforms several state of the art map registration and matching techniques by a large margin, and has a higher tolerance to large initial misalignments. We release an implementation of the proposed approach along with the acquired datasets with this paper.

\end{abstract}

\begin{IEEEkeywords}
Robotics in Agriculture and Forestry, Mapping, Multi-Robot Systems
\end{IEEEkeywords}

\section*{Supplementary Material}
\begin{quote}
  \begin{scriptsize}
    { \texttt{www.dis.uniroma1.it/{\texttildelow}labrococo/fsd/agricolmap\_sup.pdf}}
  \end{scriptsize}
\end{quote}

The datasets and our C++ implementation are available at:
\begin{quote}
  \begin{scriptsize}
    { \texttt{www.dis.uniroma1.it/{\texttildelow}labrococo/fsd}}
  \end{scriptsize}
\end{quote}


\section{Introduction}\label{sec:intro}

\IEEEPARstart{C}{ooperation} between aerial and ground robots undoubtedly offers benefits to many applications, thanks to the complementarity of the characteristics of these robots~\cite{kaslin2016collaborative}. This is especially useful in robotic systems applied to precision agriculture, where the areas of interest are usually vast. A \acrshort{uav} allows rapid inspections of large areas~\cite{khanna2015beyond}, and then share information such as crop health or weeds distribution indicators of areas of interest with an agricultural \acrshort{ugv}. The ground robot can operate for long periods of time, carry high payloads, perform targeted actions, such as fertilizer application or selective weed treatment, on the areas selected by the \acrshort{uav}. The robots can also cooperate to generate 3D maps of the environment, e.g., annotated with parameters, such as crop density and weed pressure, suitable for supporting the farmer's decision making. The \acrshort{uav} can quickly provide a coarse reconstruction of a large area, that can be updated with more detailed and higher resolution map portions generated by the \acrshort{ugv} visiting selected areas.

\begin{figure}[t!]
   \centering
   \includegraphics[width=\columnwidth]{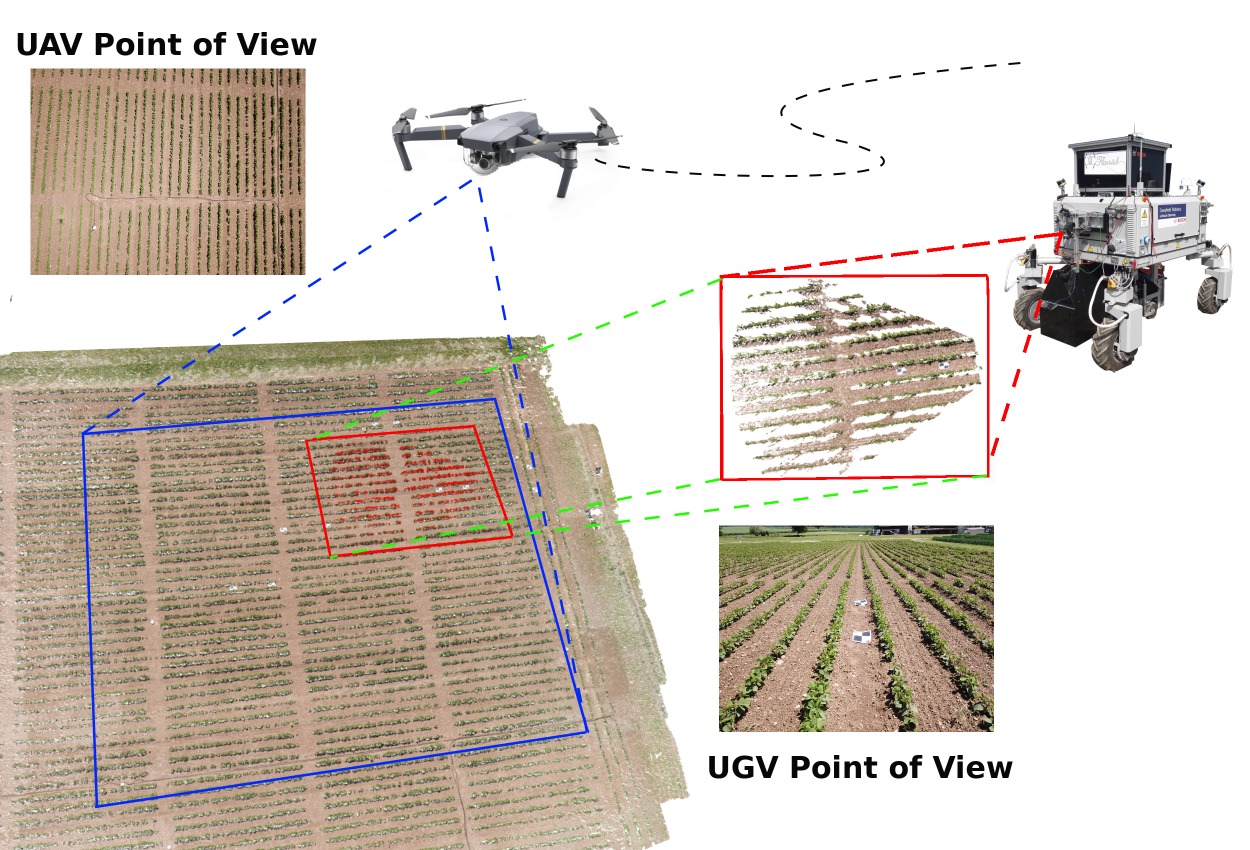}
   \caption{An overview of \ourmethod. Both the \acrshort{ugv} and \acrshort{uav} generate, using data gathered from their onboard cameras, colored point clouds of the cultivated field. The proposed method aims to accurately merge these maps by means of an affine transformation that registers the \acrshort{ugv} submap (red rectangular area) into the \acrshort{uav} aerial map (blue rectangular area), taking into account possible scale discrepancies.}
   \label{fig:motivatonal}
\end{figure}

\begin{figure}[ht!]
   \centering
   \includegraphics[width=\columnwidth]{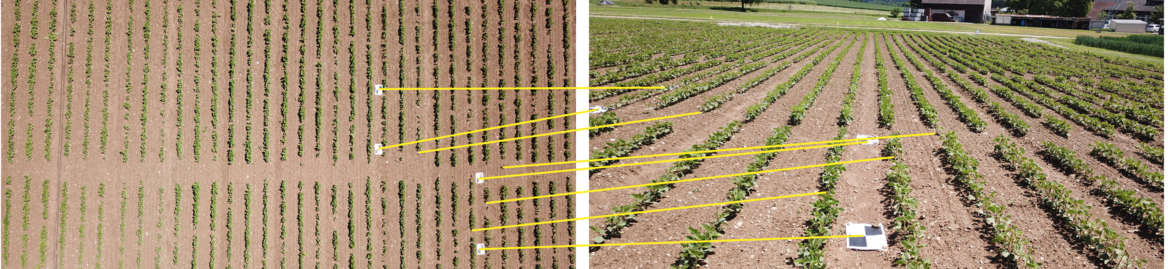}
   \caption{Pictures of the same portion of field seen from the \acrshort{uav} point-of-view (left) and from the \acrshort{ugv} point-of-view (right). The local crop arrangement geometry, such as the missing crop plants, is generally not visible from the \acrshort{ugv} point-of-view. The yellow solid lines represent an example of manually annotated correct point matches. It is important to underline the complexity required in obtaining correct data association, also from an human point-of-view. The fiducial markers on the filed have been used to compute the ground truth alignments between the maps.}
   \label{fig:data_ass_ex}
\end{figure}

All the above applications assume that both \acrshort{uav}s and \acrshort{ugv}s can share information using a unified environment model with centimeter-level accuracy, i.e. an accurate \textit{shared map} of the field. There are two classes of methods designed to generate multi-robot environment representations: (i) multi-robot \acrlong{slam} (\acrshort{slam}) algorithms (e.g., \cite{Howard2006,GIL201068}), that concurrently build a single map by fusing raw measurements or small local maps generated from multiple robots; (ii) map registration algorithms (e.g., \cite{Birk2006,Bonanni2017}) that align and merge maps independently generated by each robot into a unified map. On the one hand, the lack of distinctive visual and 3D landmarks in an agricultural field, along with the difference in the robots' point-of-views (e.g., \figref{fig:data_ass_ex}), prevent direct employment of standard multi-robot \acrshort{slam} pipelines, either based on visual or geometric features. On the other hand, merging maps independently generated by the \acrshort{uav}s and \acrshort{ugv}s in an agricultural environment is also a complex task, since maps are usually composed of similar, repetitive patterns that easily confuse conventional data association methods~\cite{gawel20173d}.
Furthermore, due to inaccuracies in the map building process, the merged maps are usually affected by local inconsistencies,  missing data, occlusions, and global deformations such as directional scale errors, that negatively affect the performance of standard alignment methods. Geolocation information associated with (i) sensor readings or (ii) maps often can't solve the limitations of conventional methods in agricultural environments, since the location and orientation accuracy provided by standard reference sensors\footnote{\acrlong{gps}s (\acrshort{gps}s) and \acrlong{ahrs}s (\acrshort{ahrs}s)} \cite{imperoli2018} is not suitable to prevent such systems from converging towards sub-optimal solutions (see \secref{sec:exp})\\

In this paper, we introduce \ourmethod, an Aerial-Ground Collaborative 3D Mapping pipeline, which provides an effective and robust solution to the cooperative mapping problem with heterogeneous robots, specifically designed for farming scenarios. We address this problem by proposing a non-rigid map registration strategy able to deal with maps with different resolutions, local inconsistencies, global deformations, and relatively large initial misalignments. We assume that both a \acrshort{uav} and a \acrshort{ugv} can generate a colored, geotagged point cloud of a target farm environment, e.g., by means of photogrammetry-based 3D reconstruction. (\figref{fig:motivatonal}). 
To solve the data association problem between the input point clouds, we propose to switch from a 3D problem to a 2D one, solved by using a global, 2D dense matching approach.
The key intuition behind this choice is that points belonging to a cloud locally share similar displacement vectors that associate such points with points in the other cloud. 
Therefore, the idea is to employ a regularized 2D matching strategy that penalizes the displacement vectors discontinuities for each point neighborhood\footnote{In other words, a regularized matching enforces the smoothness of the displacement vectors for neighboring points.}.
With this formulation, good correspondences are iteratively improved and spread through cooperative search among neighboring points.
This approach has been inspired by the \acrlong{ldof} (\acrshort{ldof}) problem in computer vision and, actually, we cast our data association problem as a \acrshort{ldof} problem. To this end, we convert the colored point clouds into a more suited, multimodal environment representation that allows one to exploit two-dimensional approaches and to highlight both the semantic and the geometric properties of the target map. The former is represented by a vegetation index map, while the latter through a \acrlong{dsm} (\acrshort{dsm}). More specifically, we transform each input point cloud into a grid representation, where each cell stores (i) the \acrlong{exg} index (\acrshort{exg}) and (ii) the local surface height information (e.g., the height of the plants, soil, etc.). Then, we use the data provided by the \acrshort{gps} and the \acrshort{ahrs} to extract an initial guess of the relative displacement and rotation between grid maps to match.
Hence, we compute a dense set of point-to-point correspondences between matched maps, exploiting a modified, state-of-the-art \acrshort{ldof} system \cite{hu2016}, tailored to the precision agriculture context. To adapt this algorithm to our environment representation, we propose to use a different cost function that involves both the ExG information and the local structure geometry around each cell. 
We select, using a voting scheme, the bigger subset of correspondences with coherent, similar flows, to be used to infer a preliminary alignment transformation between the maps. In order to deal with directional scale errors, we use a non-rigid point-set registration algorithm to estimate an affine transformation. The final registration is obtained by performing a robust point-to-point registration over the input point clouds, pruned from all points that do not belong to the vegetation. A schematic overview of the proposed approach is depicted in \figref{fig:flow_chart}.\\

We report results from an exhaustive set of experiments (\secref{sec:exp}) on data acquired by a \acrshort{uav} and a handheld camera, simulating the \acrshort{ugv}, on crop fields in Eschikon, Switzerland. We show that the proposed approach is able to guarantee with a high probability a correct registration for an initial translational error up to 5 meters, an initial heading misalignment up to 11.5 degrees, and a directional scale error of up to 30\%. We found similar registration performance across fields with three different crop species, showing that the method generalizes well across different kinds of crop species. We also report a comparison with state-of-the-art point-to-point registration and matching algorithms, showing that our approach outperforms them in all the experiments. 

\subsection{Related Work}
The field of multi-robot cooperative mapping is a recurrent and relevant problem in literature and, as previously introduced, several solutions have been presented by means of either multi-robot \acrshort{slam} algorithms or map merging/map registration strategies, in both 2D (\cite{Birk2006,Blanco2013,Saeedi2011}) and 3D (\cite{Bonanni2017,FruehCVPR2003,Jessup2014}) settings. Registration of point cloud based maps can also be considered as an instance of the more general point set registration problem \cite{Chui2003,Fitzgibbon01c}. In this work, we mainly review methods based on map registration, since the heterogeneity of the involved robots and the lack of distinctive visual and geometrical features on an agricultural environment prevent the employment of standard multi-robot \acrshort{slam} methods; a comprehensive literature review about this class of methods can be found in \cite{doi:10.1002/rob.21620}.\\
Map registration is a challenging problem especially when dealing with heterogeneous robots, where data is gathered from different points-of-view and with different noise characteristics. It has been intensively investigated, especially in the context of urban reconstruction with aerial and ground data. In~\cite{shan2014}, the authors focus on the problem of geo-registering ground-based multi-view stereo models by proposing a novel viewpoint-dependent matching method. Wang \emph{et al.}~\cite{wang2013} deal with aligning 3D structure-from-motion point clouds obtained from Internet imagery with existing geographic information sources, such as noisy geotags from input Flickr photos and geotagged city models and images collected from Google Street View and Google Earth. B\'odis-Szomor\'u \emph{et al.}~\cite{bodis2016} propose to merge low detailed airborne point clouds with incomplete street-side point clouds by applying volumetric fusion based on a 3D tetrahedralization (3DT). Fr\"uh \emph{et al.}~\cite{fruh2003} propose to use \acrlong{dsm}s obtained from a laser airborne reconstruction to localize a ground vehicle equipped with 2D laser scanners and a digital camera, detailed ground-based facade models are hence merged with a complementary airborne model. Michael \emph{et al.} \cite{Michael2012} propose a collaborative \acrshort{uav}-\acrshort{ugv} mapping approach in earthquake-damaged contexts. They merge the point clouds generated by the two robots using a 3D \acrlong{icp} (\acrshort{icp}) algorithm, with an initial guess provided by the (known) \acrshort{uav} takeoff location; the authors make the assumption that the environment is generally described by flat planes and vertical walls, also called the ``Manhattan world'' assumption. The \acrshort{icp} algorithm has also been exploited in \cite{Forster2013} and \cite{Hinzmann2016}. Forster \emph{et al.}~\cite{Forster2013} align dense 3D maps obtained by a \acrshort{ugv} equipped with an RGB-D camera and by a \acrshort{uav} running dense monocular reconstruction: they obtain the initial guess alignment between the maps by localizing the \acrshort{uav} with respect to the \acrshort{ugv} with a Monte Carlo Localization method applied to height-maps computed by the two robots. Hinzmann \emph{et al.} \cite{Hinzmann2016} deal with the registration of dense \acrshort{lidar}-based point clouds with sparse image-based point clouds by proposing a probabilistic data association approach that specifically takes the individual cloud densities into consideration. In \cite{Gawel2016}, Gawel \emph{et al.} present a registration procedure for matching \acrshort{lidar} point-cloud maps and sparse vision keypoint maps by using structural descriptors. \\
Although much literature addresses the problem of map registration for heterogeneous robots, most of the proposed methods make strong context-based assumptions, such as the presence of structural or visual landmarks, ``Manhattan world'' assumptions, etc. Registering 3D maps in an agricultural setting, in some respects, is even more challenging: the environment is homogeneous, poorly structured and it usually gives rise to strong sensor aliasing. For these reasons, most of the approaches mentioned above cannot directly be applied to an agricultural scenario. Localization and mapping in an agricultural scenario is a  topic that is recently gathering great attention in the robotics community \cite{weiss2011plant,english2014,imperoli2018}. 
Most of these systems, however, deal with a single robot, and the problem of fusing maps built from multiple robots is usually not adequately addressed and a little, very recent research exists on this topic. Dong \emph{et al.} \cite{dong2017} propose a spatio-temporal reconstruction framework for precision agriculture that aims to merge multiple 3D field reconstructions of the same field across time. They use single row reconstructions as starting points for the data association, that is actually performed by using standard visual features. This method uses images acquired by a single \acrshort{ugv} that moves in the same field at different times and, being based on visual features, cannot manage drastic viewpoint changes or large misalignments when matching aerial and ground maps. A local feature descriptor designed to deal with large viewpoint changes has been proposed by Chebrolu \emph{et al.} in \cite{Chebrolu2018}. The authors propose to encode with such descriptor the almost static geometry of the crop arrangement in the field.
Despite the promising results, this method suffers from the presence of occluded areas when switching from the \acrshort{uav} to the \acrshort{ugv} point-of-view.

\subsection{Contributions}

Our contributions are the following: (i) A map registration framework specifically designed for heterogeneous robots in an agricultural environment; (ii) To the best of our knowledge, we are the first to apply a \acrshort{ldof} based 3D map alignment; (iii) Extensive performance evaluations that show the effectiveness of our approach; (iv) An open-source implementation of our method and three challenging datasets with different crop species with ground truth.

\section{Problem Statement and Assumptions}\label{sec:problem}

\begin{figure*}[ht!]
   \centering
   \includegraphics[width=.98\linewidth]{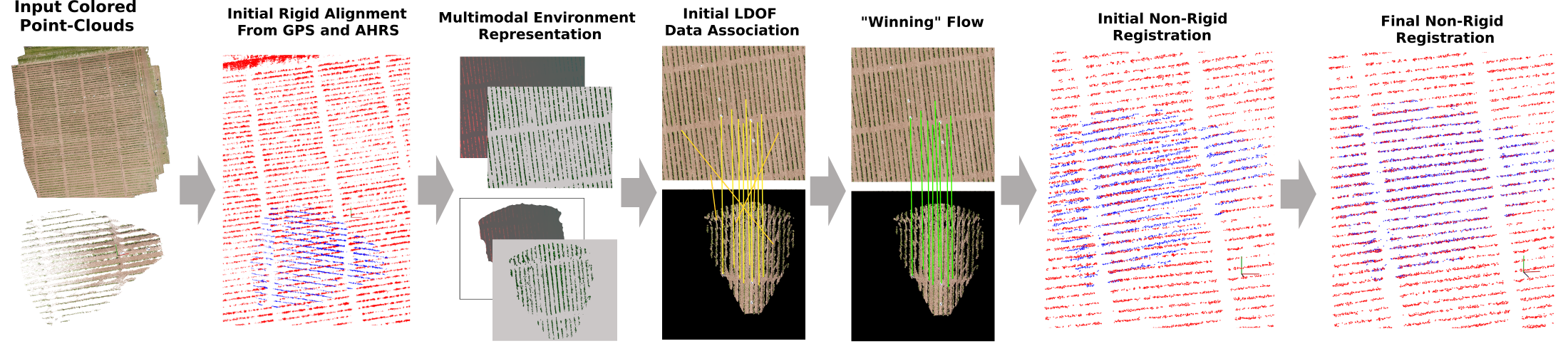}
   \caption{Overview of the proposed approach. For visualization purposes, in column 2,7 and 8 we colored in blue and red the \acrshort{ugv} and \acrshort{uav} point clouds, respectively, pruned from all points that do not belong to vegetation, according to a thresholding operator applied to the \acrshort{exg} index. Starting from the left side, we show: (i) the input colored point clouds gathered by the \acrshort{uav} and \acrshort{ugv}; (ii) the initial noisy and biased rigid alignment provided by the \acrshort{gps} and the \acrshort{ahrs}; (iii) the generated multimodal grid maps; (iv) the initial \acrshort{ldof} data associations, i.e. the point-to-point correspondences, in yellow; (v) the ''winning`` data associations (flows), in green, selected by a voting scheme; (vi) the aligned point clouds according to the initial affine transform; (vii) the final non-rigid registration after the refinement step.}
   \label{fig:flow_chart}
\end{figure*}

Given two 3D colored point clouds $\mathcal{M}_A$ and $\mathcal{M}_G$ of a farmland (\figref{fig:flow_chart}, first column), built from data gathered from a \acrshort{uav} and a \acrshort{ugv}, respectively, our goal is to find a transformation $F:\mathbb{R}^3 \rightarrow \mathbb{R}^3$ that allows to accurately align them. 
$\mathcal{M}_A$ and $\mathcal{M}_G$ can be generated, for instance, by using an off-the-shelf photogrammetry-based 3D reconstruction software applied to sequences of geotagged images. Our method makes the following assumptions:

\begin{enumerate}
 \item The input maps built form \acrshort{uav}s and \acrshort{ugv}s data can have different spatial resolutions but they refer to the same field, with some overlap among them;
 \item The data used to build the maps were acquired at approximately the same time;
 \item The maps are roughly geotagged, possibly with noisy locations and orientations;
 \item They can be affected by local inconsistencies, missing data, and deformations, such as directional scale errors.
 \item $\mathcal{M}_A$ is not affected by any scale inconsistencies.
\end{enumerate}
Hypotheses 1, 2, and 3 are the essential data requirements. Hypothesis 4) implies the violation of the typical rigid-body transformation assumption between the two maps: therefore, we represent $F$ as an affine transformation that allows anisotropic (i.e., non-uniform) scaling between the maps. Hypothesis 5) is an acceptable assumption, since the map created by the \acrshort{uav} is usually wider than $\mathcal{M}_G$, and generated by using less noisy GPS readings, so the scale drift effect tends to be canceled: hence, we look for a transformation that aligns $\mathcal{M}_G$ with $\mathcal{M}_A$ by correcting the scale errors of $\mathcal{M}_G$ with respect to $\mathcal{M}_A$.

\section{Data Association}\label{sec:data_association}

In order to estimate the transformation $F$ that aligns the two maps, we need to find a set of point correspondences, $m_{A,G} = \{(p, q) : p \in \mathcal{M}_A, q \in \mathcal{M}_G\}$ between $\mathcal{M}_A$ and $\mathcal{M}_G$, that represent points pairs belonging to the same global 3D position. As introduced before and shown in the experiments (see \secref{sec:exp}), conventional sparse matching approaches based on local descriptors are unlikely to provide effective results due to the big amount of repetitive and non-distinctive patterns spread over farmlands. Instead, inspired by the fact that when the maps are misaligned, points in $\mathcal{M}_A$ locally share a coherent ''flow`` towards corresponding points in $\mathcal{M}_G$, our method casts the data association estimation problem as a \textit{dense}, \textit{regularized}, matching approach. This problem resembles the dense optical flow estimation problem for RGB images: in this context, global methods (e.g., \cite{Horn1981}) aim to build correspondences pixel by pixel between a pair of images by minimizing a cost function that, for each pixel, involves a data term that measures the point-wise similarity and a regularization term that fosters smoothness between nearby flows (i.e., nearby pixel to pixel associations).

\subsection{Multimodal Grid Map}\label{sec:multimodal_grid}

Our goal is to estimate $m_{A,G}$ by computing a ''dense flow`` that, given an initial, noisy alignment between the maps provided by a \acrshort{gps} and a \acrshort{ahrs} (\figref{fig:flow_chart}, second column), associates points in $\mathcal{M}_A$  with points in $\mathcal{M}_G$. Unfortunately, conventional methods designed for RGB images are not directly applicable to colored point clouds: we introduce here a multimodal environment representation that allows to exploit such methods while enhancing both the semantic and the geometrical properties of the target map. A cultivated field is basically a globally flat surface populated by plants. A \acrshort{dsm}\footnote{A \acrshort{dsm} is a raster representations of the height of the objects on a surface.} can well approximate the field structure geometry, while a vegetation index can highlight the meaningful parts of the field and the visual relevant patterns: in our environment representation, we exploit both these intuitions. We generate a \acrshort{dsm} from the point cloud; for each cell of the \acrshort{dsm} grid, we also provide an \acrshort{exg} index that, starting from the RGB values, highlights the amount of vegetation. More specifically, we transform a colored point cloud $\mathcal{M}$ into a two dimensional grid map $\mathcal{J}:\mathbb{R}^2 \rightarrow \mathbb{R}^2$ (\figref{fig:flow_chart}, third column), where for each cell we provide the surface height and the \acrshort{exg} index, with the following procedure:
\begin{enumerate}
 \item We select a rectangle that bounds the target area by means of minimum-maximum latitude and longitude;
 \item The selected area is discretized into a grid map $\mathcal{J}$ of $w \times h$ cells, by using a step of $s$ meters. In practice, each of the $w \times h$ cells represents a square of $s \times s$ meters. Each cell is initialized with $(0,0)$ pairs.
 \item Remembering that $\mathcal{M}$  is geotagged (see \secref{sec:problem}), we can associate each 3D point of $\mathcal{M}$ to one cell of $\mathcal{J}$ just using the $x,y$ and yaw information.
 \item For each cell with associated at least one 3D point: (a) We compute the height as a \textit{weighted} average of the $z$ coordinates of the 3D points that belong to such cell; (b) We compute the \acrshort{exg} index as a \textit{weighted} average of the \acrshort{exg} indexes of the 3D points that belong to such cell, where for each point $p$ we have:
      \begin{equation}
         \acrshort{exg}(p) = 2g_p - r_p - b_p.
     \end{equation}
  with $r_p$, $g_p$ and $b_p$ the RGB components of the point; (c) we store the 3D global position of the nearest point in the original colored point cloud.
\end{enumerate}
  
Both the averages use as weighting factor a circular, bivariate Gaussian distribution with standard deviation $\sigma_{avg}$: points with $x,y$ coordinates close to center of the cell get a higher weight.

\subsection{Multimodal \acrlong{ldof}}\label{sec:ldof}
We generate from both the $\mathcal{M}_A$ and $\mathcal{M}_G$ the corresponding multimodal representations $\mathcal{J}_A$ and $\mathcal{J}_G$. In the ideal case, with perfect geotags and no map deformations, a simple geotagged superimposition of the two maps should provide a perfect alignment: the ''flow`` that associates cells between the two maps should be zero. Unfortunately, in the real case, due to the inaccuracies of both the geotags and the 3D reconstruction, non zero, potentially large displacements are introduced in the associations. These offsets are locally consistent but not constant for each cell, due to reconstruction errors. To estimate the offsets map, we employ a modified version of the \acrlong{cpm} (\acrshort{cpm}) framework described in \cite{hu2016}. \acrshort{cpm} is a recent \acrshort{ldof} system that provides cutting edge estimation results even in presence of very large displacements, and is more efficient than other state-of-the-art methods with similar accuracy.\\
For efficiency, \acrshort{cpm} looks for the best correspondences of some seeds that are refined by means of a dense, iterative neighborhood propagation: the seeds are a set of points regularly distributed within the image. Given two images $\mathcal{I}_0,\mathcal{I}_1 \in \mathbb{R}^2$ and a collection of seeds $S = \{ s_1, \dots, s_n \}$ at position $\{p(s_1),\dots,p(s_n)\}$, the goal of this framework is to determine the flow of each seed $f(s_i) = M(p(s_i))-p(s_i) \in \mathbb{R}^2$, where $M(p(s_i))$ is the corresponding matching position in $\mathcal{I}_1$ for the seed $s_i$ in $\mathcal{I}_0$. The flow computation for each \textit{seed} is  performed by an iterative, coarse-to-fine random search strategy that minimizes a cost function:
\begin{equation}
f(s_i) = \underset{f_{s_j}}{\operatorname{argmin}}(C(f(s_j))), s_j \in s_i \cap \mathcal{N}_i\label{eq:patchmatch}
\end{equation}
where $C(f(\cdot))$ denotes the match cost between the patch centered at $p(s_i)$ in $\mathcal{I}_0$ and the patch centered in $p(s_i)+f(\cdot)$ in $\mathcal{I}_1$, while $\mathcal{N}_i$  is a set of spatially adjacent neighbors seeds around $s_i$ whose flow has already been computed in the current iteration with Eq.~\ref{eq:patchmatch}.
For a comprehensive description of the flow estimation pipeline, we refer the reader to \cite{hu2016}. 

Our goal is to use the \acrshort{cpm} algorithm to compute the flow between $\mathcal{J}_A$ and $\mathcal{J}_G$. To exploit the full information provided by our grid maps (see \secref{sec:multimodal_grid}), we modified the \acrshort{cpm} matching cost in order to take into account both the height and \acrshort{exg} channels. We split the cost function in two terms:
\begin{equation}
\label{eq:cost_function}
C_{flow}(f(s_i)) = \alpha \cdot C_{DY}(f(s_i)) + \beta \cdot C_{\acrshort{fpfh}}(f(s_i))
\end{equation}

$C_{DY}(f(s_i))$ is the DAISY~\cite{10.1109/TPAMI.2009.77} based match cost as in the original \acrshort{cpm} algorithm: in our case the DAISY descriptors have been computed from the \acrshort{exg} channel of $\mathcal{J}_A$ and $\mathcal{J}_G$. $C_{\acrshort{fpfh}}(f(s_i))$ is a match cost computed using the height channel. 
We chose the \acrlong{fpfh} (\acrshort{fpfh})~\cite{rusu2009} descriptor for this second term: the \acrshort{fpfh} descriptors are robust multi-dimensional features
which describe the \textit{local} geometry of a point cloud\footnote{It is noteworthy to highlight that the FPFH, being a local descriptor, does not embed global displacements along the axes.}, in our case they are computed from the organized point cloud generated from the height channel of $\mathcal{J}_A$ and $\mathcal{J}_G$. The parameters $\alpha$ and $\beta$ are the weighting factors of the two terms. As in \cite{hu2016}, the patch-based matching cost is chosen to be the sum of the absolute difference over all the 128 and 32 dimensions of the DAISY and \acrshort{fpfh} flows, respectively, at the matching points. The proposed cost function takes into account both the visual appearance and the local 3D structure of the plants.\\
Once we have computed the dense flow between $\mathcal{J}_A$ and $\mathcal{J}_G$ (\figref{fig:flow_chart}, fourth column), we extract the largest set of coherent flows by employing a voting scheme inspired by the classical Hough transform with discretization step $t_f$; these flows define a set of point-to-point matches $m_{A,G}$ that will be used to infer a preliminary alignment (\figref{fig:flow_chart}, fifth column). 

\section{Non-Rigid Registration}

\begin{figure*}[ht]
\centering
\begin{minipage}[b]{0.5\columnwidth}
\includegraphics[width=\linewidth]{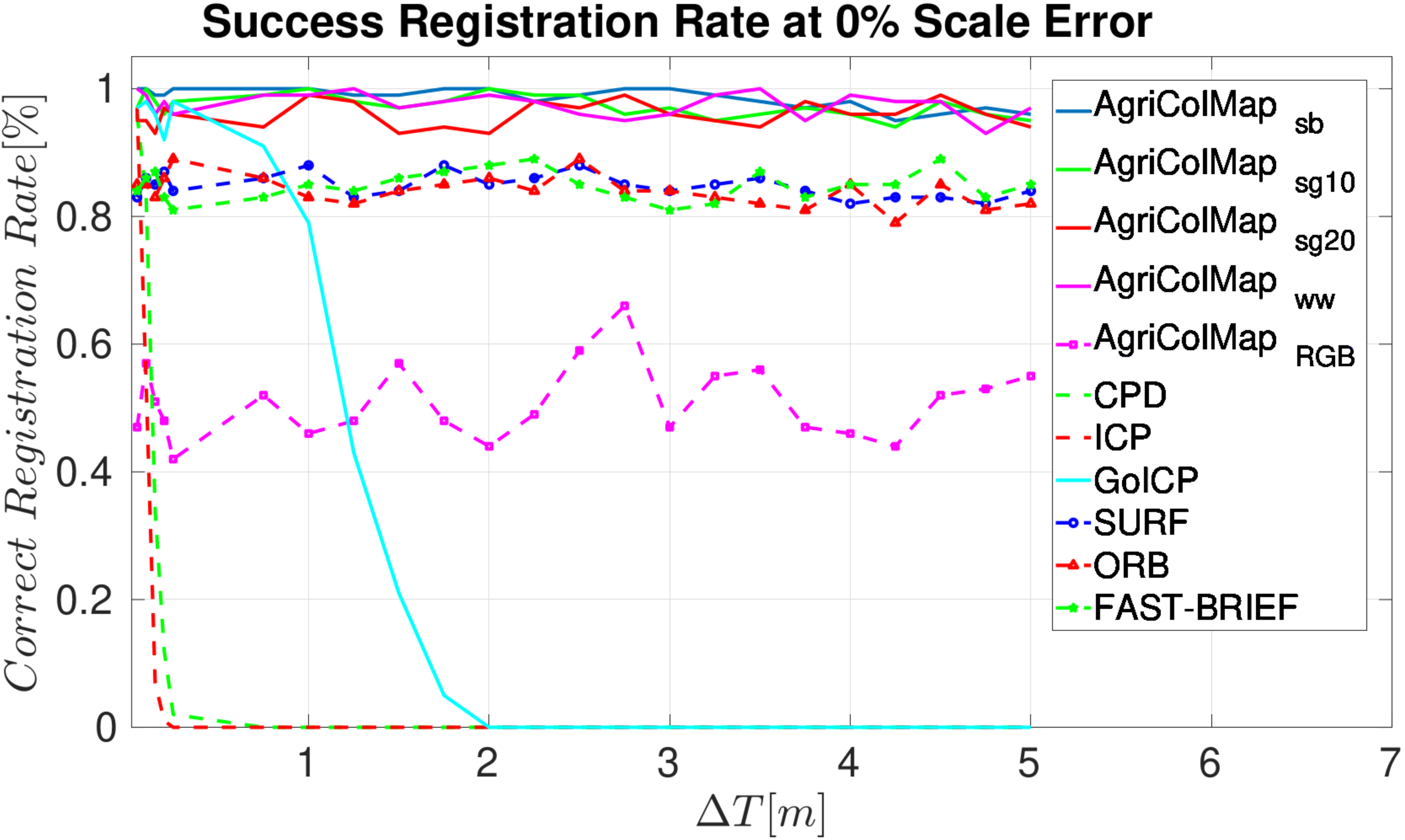}
\end{minipage}\hfill
\begin{minipage}[b]{0.5\columnwidth}
\includegraphics[width=\linewidth]{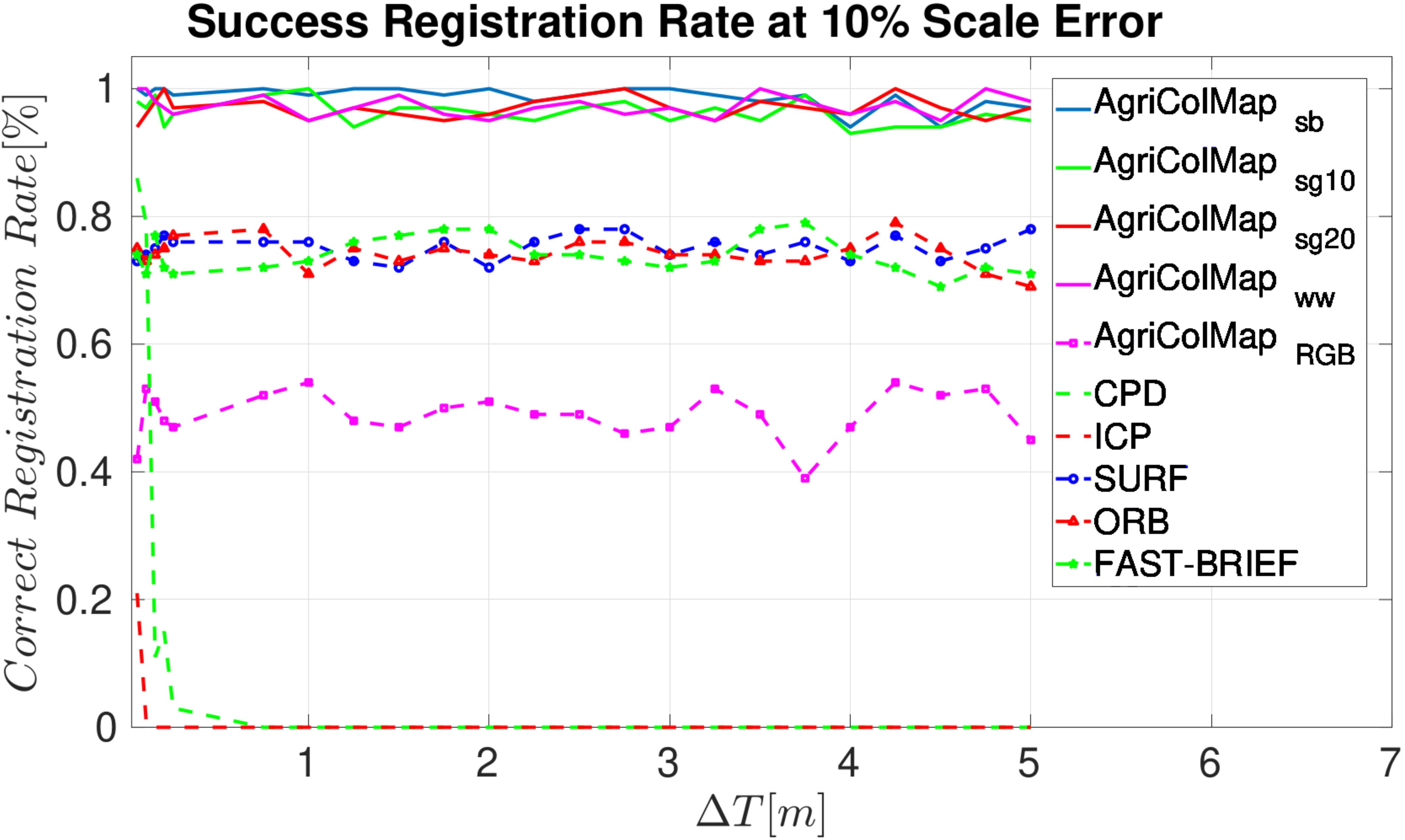}
\end{minipage}\hfill
\begin{minipage}[b]{0.5\columnwidth}
\includegraphics[width=\linewidth]{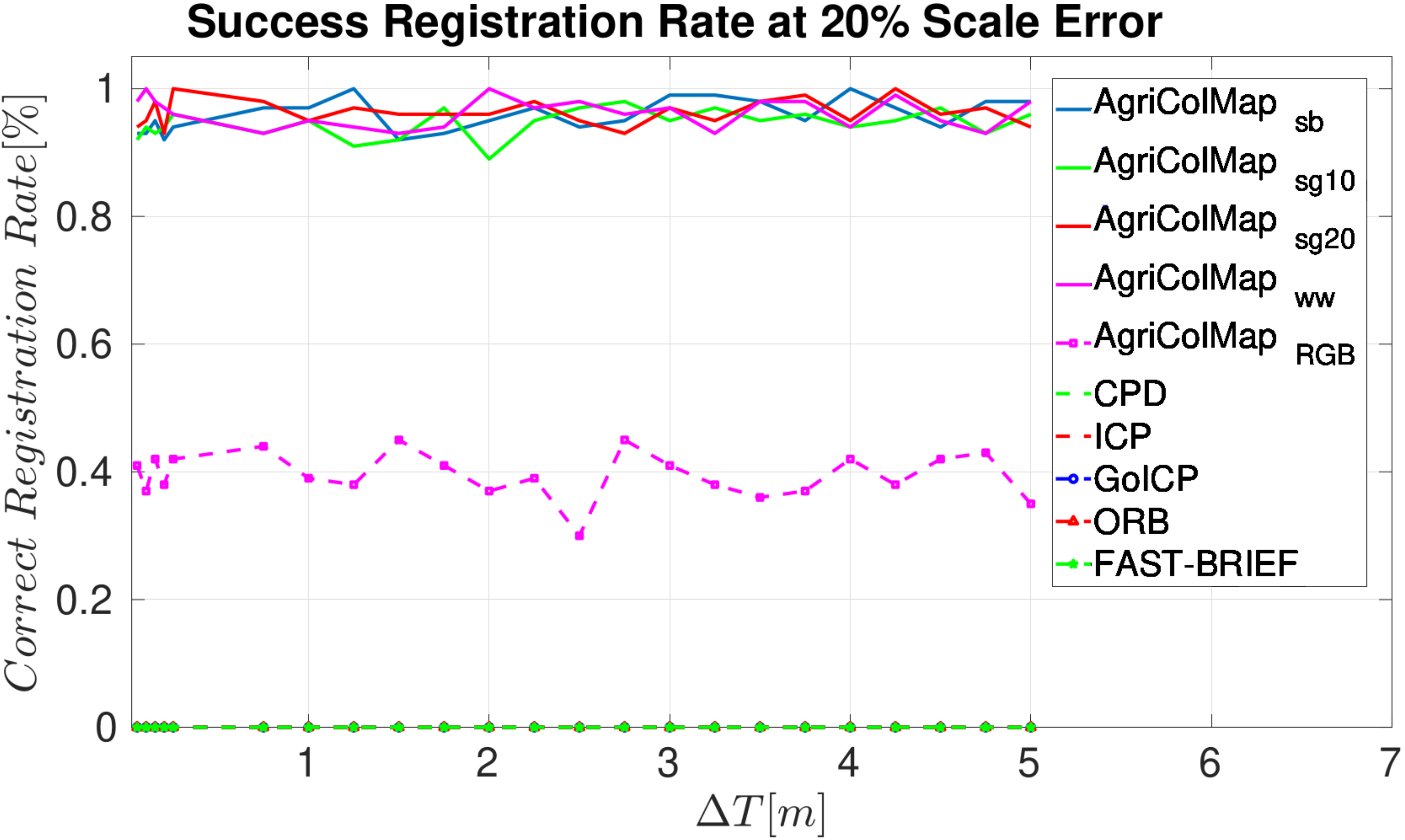}
\end{minipage}\hfill\
\begin{minipage}[b]{0.5\columnwidth}
\includegraphics[width=\linewidth]{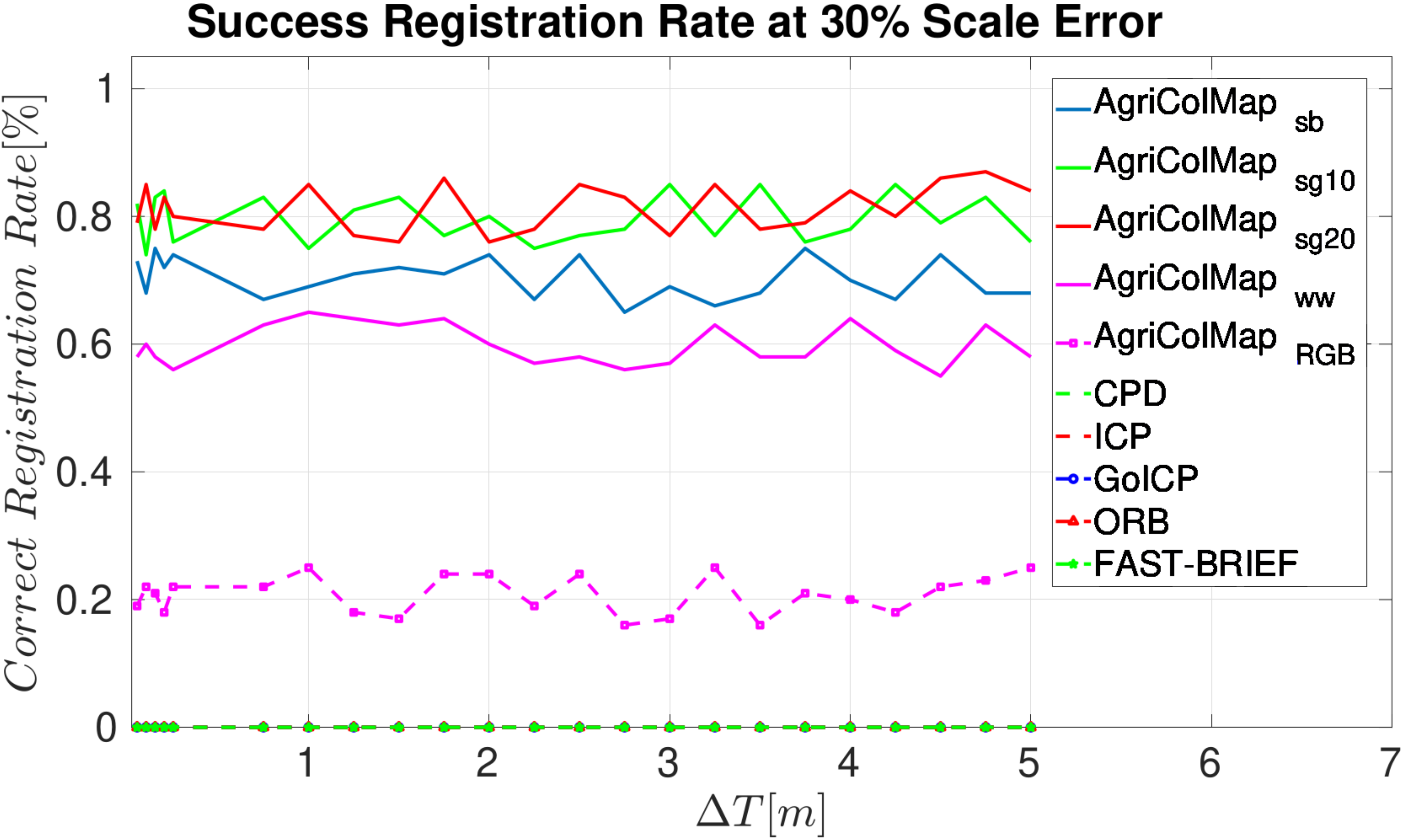}
\end{minipage}\hfill\\
\begin{minipage}[b]{0.5\columnwidth}
\includegraphics[width=\linewidth]{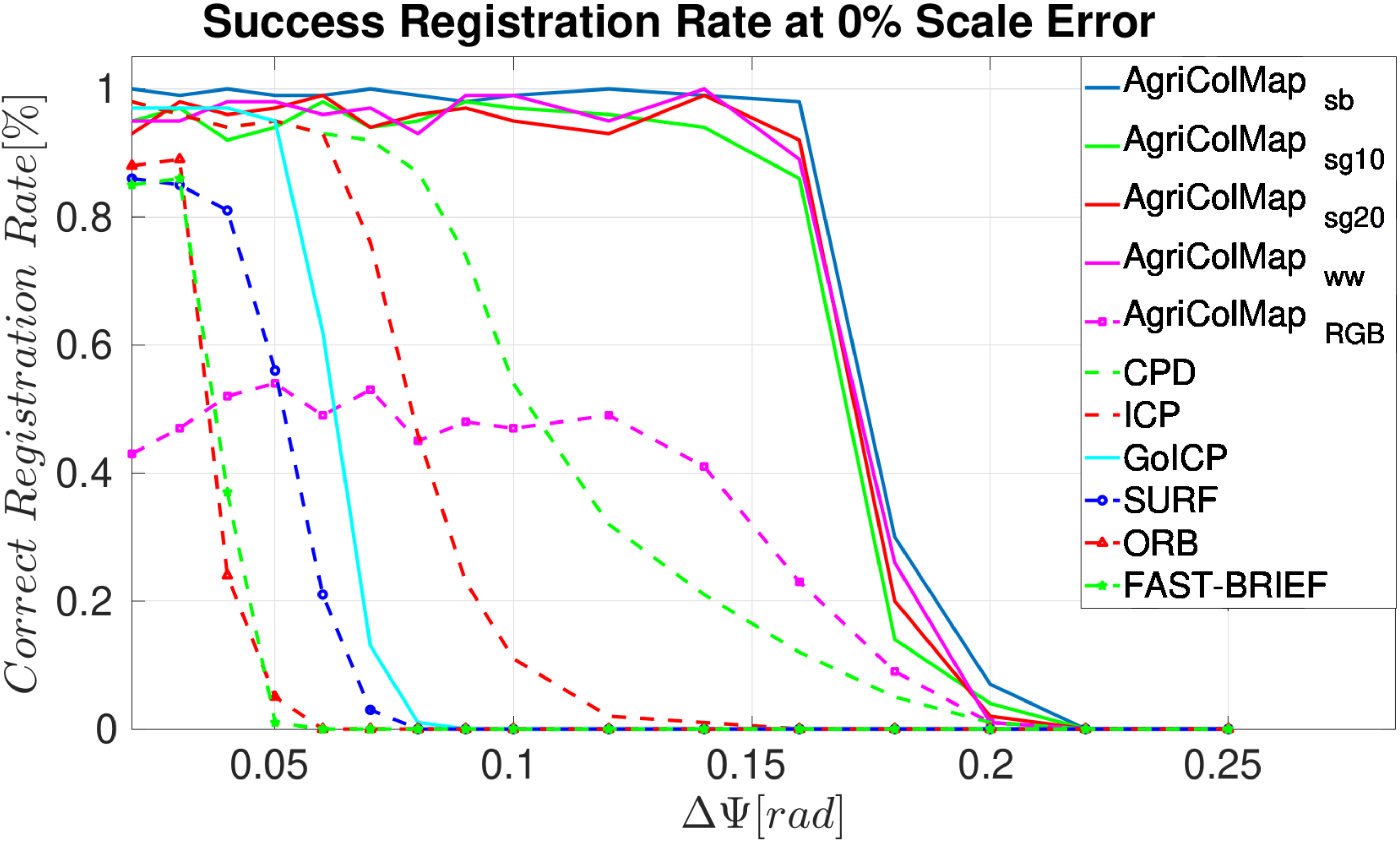}
\end{minipage}\hfill
\begin{minipage}[b]{0.5\columnwidth}
\includegraphics[width=\linewidth]{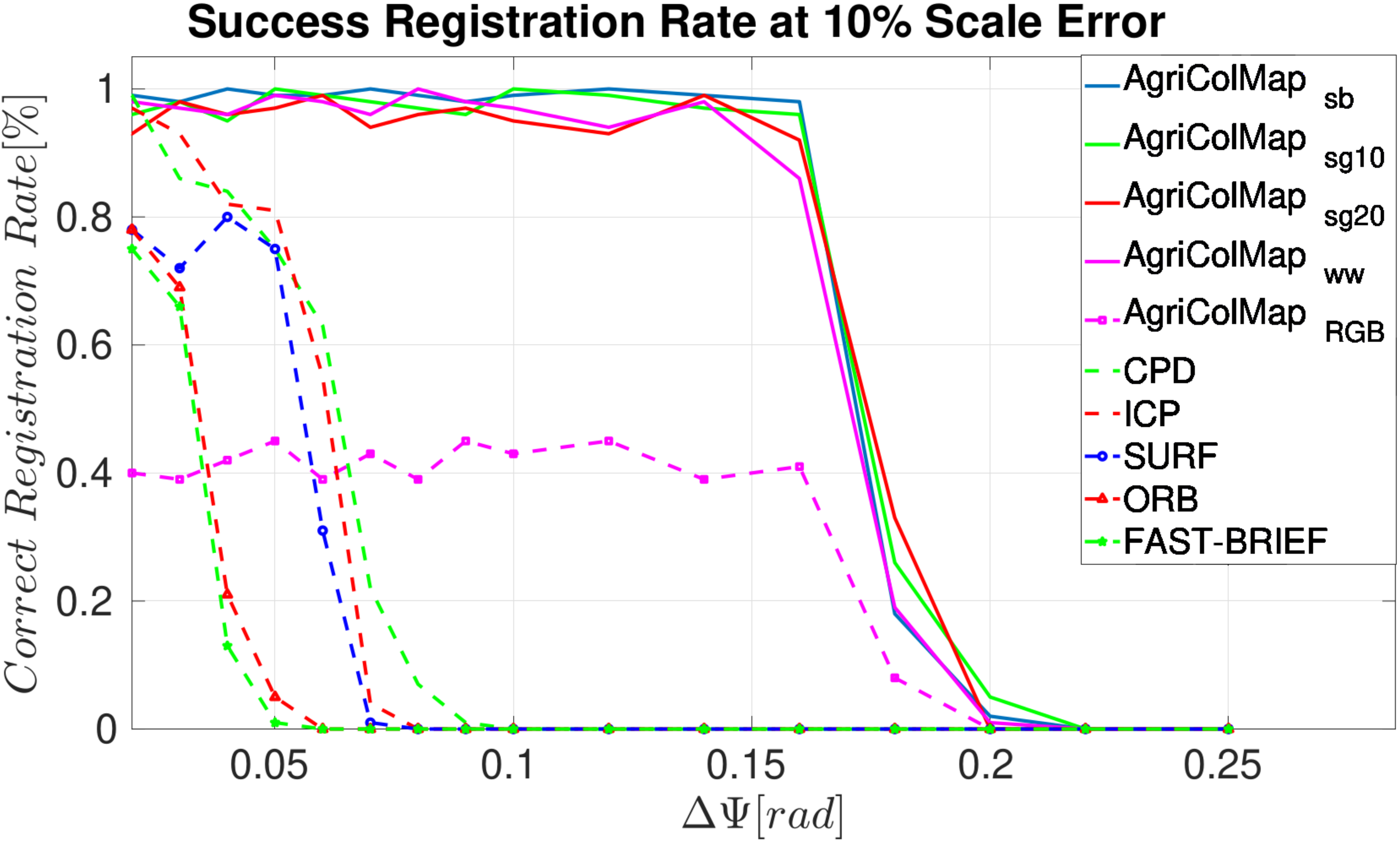}
\end{minipage}\hfill
\begin{minipage}[b]{0.5\columnwidth}
\includegraphics[width=\linewidth]{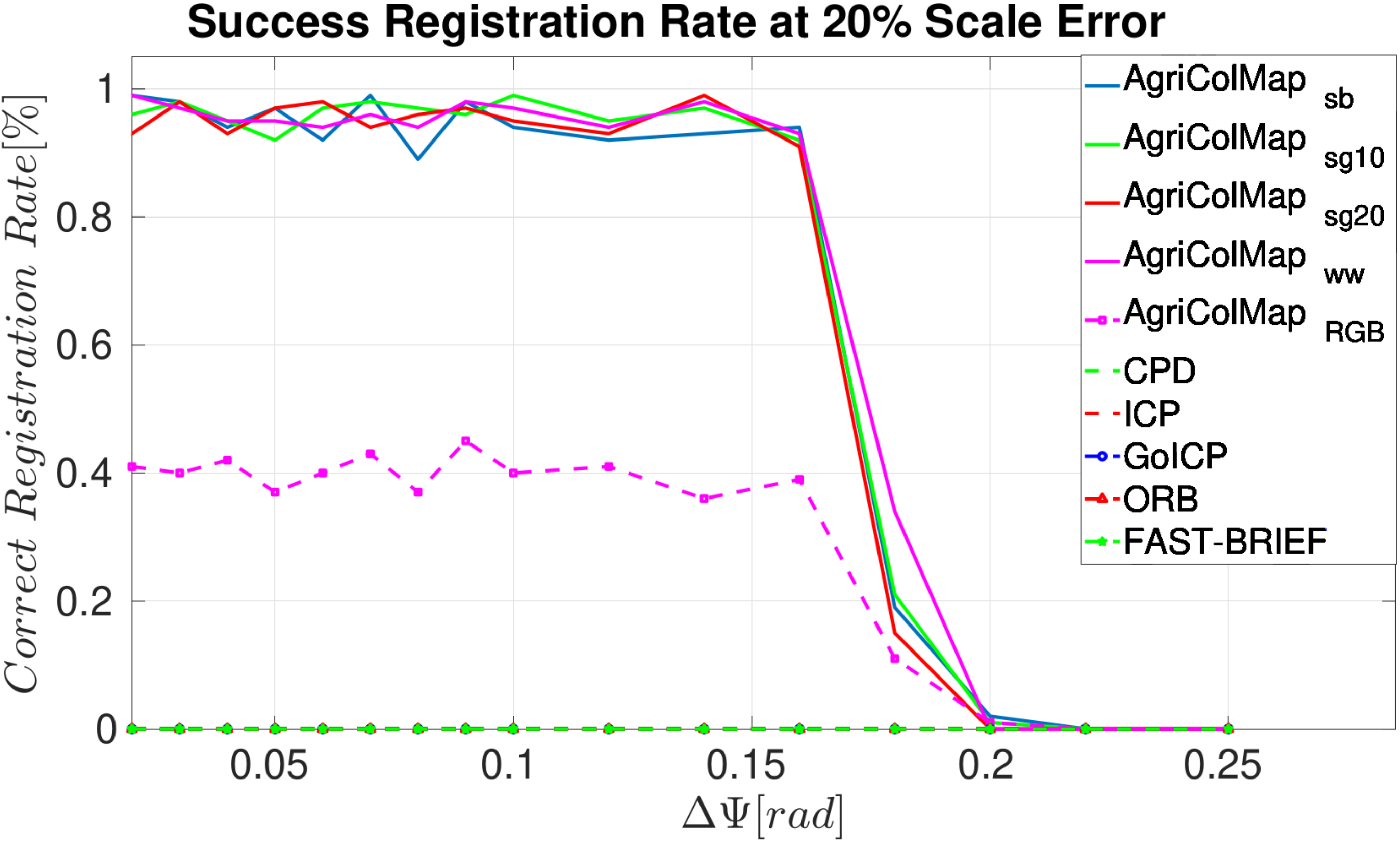}
\end{minipage}\hfill
\begin{minipage}[b]{0.5\columnwidth}
\includegraphics[width=\linewidth]{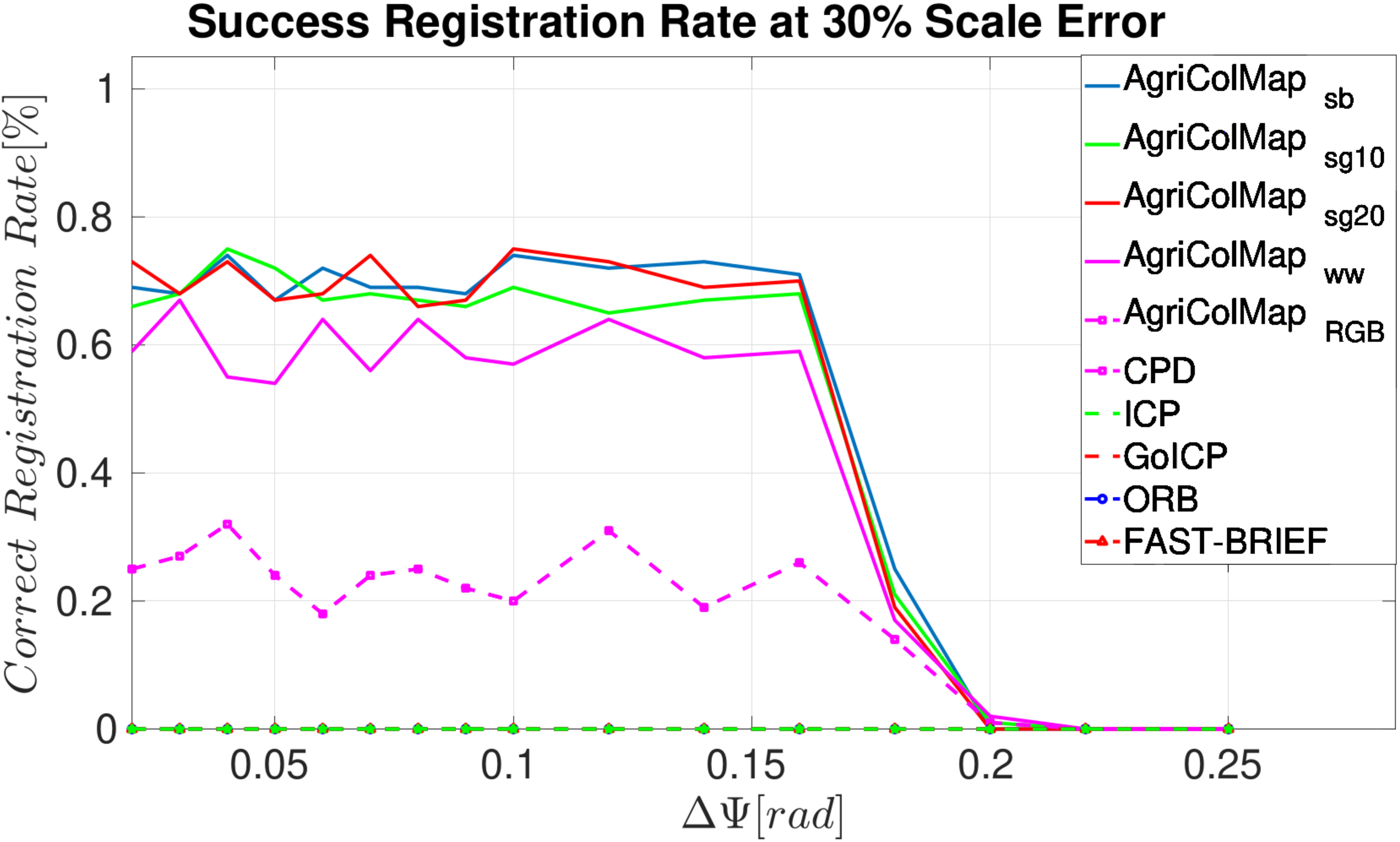}
\end{minipage}\hfill
\caption{Average success registration rate curves by varying the initial guess and the initial scale error: (i) from left to right, the initial scale error is incrementally increased:  $0\%, 10\%, 20\%, 30\%$; (ii) in each plot within the upper row, the initial heading error $\delta \psi$ is kept fixed, while the initial translational misalignment $\delta t$ is incrementally increased until 5 meters. We assume 5 meters to be a reasonable upper bound for the initial GPS translational error. (iii) in the lower row plots, $\delta \psi$ is incrementally increased, while the initial translational misalignment $\delta t$  is kept constant. It is important to point out that the successful registration rate of the \acrshort{goicp} \cite{yang2016} method is only reported for the cases without an initial scale error since this approach only deals with rigid transformations. For \ourmethod, we report the different results obtained in each dataset (sb: Soybean, sg10: Sugar Beet 10m, sg20: Sugar Beet 20m, ww: Winter Wheat).}   
\label{fig:correct_reg_rate}
\end{figure*}

The estimation of the non-rigid transformation between the maps is addressed in two steps. A preliminary affine transformation $\hat{F}$ is computed by solving a non-rigid registration problem with known point-to-point correspondences. We compute $\hat{F} = (\hat{s}\hat{R}|\hat{t})$ by solving an optimization problem with cost function the sum of the squared distances between corresponding points (\figref{fig:flow_chart}, sixth column):
\begin{equation}
C_{reg}(\hat{F}) = \sum_{i=0}^{N}{||p_i-\hat{s}\hat{R}q_i - \hat{t}||^2}
\label{eq:non_rigid_obj_fn}
\end{equation}
with $(p_i, q_i) \in m_{A,G}$, $N$ the cardinality of $m_{A,G}$, $\hat{R}$ and $\hat{t}$ the rotation matrix and the translation vector, and $\hat{s}$ is a scaling vector.	
To estimate the final registration, we firstly select from the input colored point clouds $\mathcal{M}_A$ and $\mathcal{M}_G$ two subsets, $\mathcal{M}^{veg}_A$ and $\mathcal{M}^{veg}_G$, that include only points that belong to vegetation. The selection is performed by using an \acrshort{exg} based thresholding operator over $\mathcal{M}_A$ and $\mathcal{M}_G$. This operation enhances the morphological information of the vegetation, while reducing the size of the point clouds to be registered. We finally estimate the target affine transformation $F$ by exploiting the \acrlong{cpd} (\acrshort{cpd})~\cite{myronenko2010} point set registration algorithm over the point clouds $\mathcal{M}^{veg}_A$ and $\mathcal{M}^{veg}_G$, using $\hat{F}$ as initial guess transformation.

\section{Experiments}
\label{sec:exp}

In order to analyze the performance of our system, we acquired datasets on fields of 3 different crop types in Eschikon (Switzerland) - soybean, sugar beet, and winter wheat. For each crop species we collected: (i) one sequence of GPS-IMU tagged images over the entire field from a \acrshort{uav} flying at 10 meters altitude; (ii) 4-6 sequences of GPS/IMU-tagged images of small portions of the field from a \acrshort{ugv} point-of-view. Additionally, for the sugar beet field, we acquired an additional aerial sequence of images from 20 meters altitude. More comprehensive details regarding the acquired datasets are reported in Table~\ref{tab:dataset_overview}.

The \acrshort{uav} datasets were acquired using a DJI Mavic Pro \acrshort{uav} equipped with a 12 MP color camera, while the \acrshort{ugv} datasets were acquired moving the same camera by hand with a forward-looking point-of-view, simulating data acquisition by a ground robot. The collected images are first converted into 3D colored point clouds using Pix4Dmapper~\cite{pix4d}, a professional photogrammetry software suite, which are then aligned using the proposed registration approach. To analyze the performance of the proposed approach, we make use of the following error metrics:
\begin{align}
\delta t = t-\widetilde{t} \hspace{.5cm}
\delta r &= Trace(R^T\cdot\widetilde{R}) \hspace{.5cm}
\delta s = s\oslash\widetilde{s} \\
e_t = \norm{\delta t}_2\hspace{.2cm} 
e_R &= acos( (\delta r-1)/2 )\hspace{.2cm} 
e_s = \norm{\delta s}_2
\label{eq:metrics}
\end{align}

where $\oslash$ stands for the element-wise division operator and $(e_t, e_r, e_s)$ are, respectively, the translational, the rotational, and the scale error metrics. We report the \ourmethod~related parameters we used in all the experiments in \tabref{tab:parameters}.

\begin{table}[ht!]
	\scriptsize
	\centering
	\setlength{\tabcolsep}{4pt} 
	\caption{Overview of the Datasets: the global scale error is, in general, bigger in the \acrshort{ugv} datasets since the camera is carried by hand, and therefore some \acrshort{gps} satellite signals might be not received. }
	\label{tab:dataset_overview}
	  \begin{tabular}{ cccccc }
	  \makecell{Crop Type} &Name &\makecell{\# Images} &\makecell{Crop Size \\ (avg.)} &\makecell{Global Scale \\Error} &\makecell{Recording \\Height (approx.)} \\\cline{1-6}\noalign{\vskip 1mm}

	  \multirow{4}{*}{\parbox[c]{1cm}{\centering Soybean}}

	  &\textit{s\acrshort{ugv} A}  &16  &6 cm  &$4\%$ &1 m \\
	  &\textit{s\acrshort{ugv} B}  &19  &6 cm  &$6\%$ &1 m \\
	  &\textit{s\acrshort{ugv} C}  &22  &6 cm  &$7\%$ &1 m \\
	  &\textit{s\acrshort{uav}}    &89  &6 cm  &$3\%$ &10 m \\
	  \cline{1-6}\noalign{\vskip 1mm}
	  
      \multirow{5}{*}{\parbox[c]{1cm}{\centering Sugar Beet}}

	  &\textit{sb\acrshort{ugv} A} &25  &5 cm  &$6\%$ &1 m \\
	  &\textit{sb\acrshort{ugv} B} &26  &5 cm  &$7\%$ &1 m \\
	  &\textit{sb\acrshort{ugv} C} &27  &5 cm  &$5\%$ &1 m \\
	  &\textit{sb\acrshort{uav} A} &213 &5 cm  &$3\%$ &10 m \\
	  &\textit{sb\acrshort{uav} B} &96  &5 cm  &$2\%$ &20 m \\
	  \cline{1-6}\noalign{\vskip 1mm}
	  
      \multirow{3}{*}{\parbox[c]{1cm}{\centering Winter Wheat}}

	  &\textit{ww\acrshort{ugv} A} &59  &25 cm &$9\%$ &1 m \\
	  &\textit{ww\acrshort{ugv} B} &61  &25 cm &$9\%$ &1 m \\
	  &\textit{ww\acrshort{uav}}   &108 &25 cm &$5\%$ &10 m \\
	  
	  \cline{1-6} 
	  \end{tabular}	
\end{table}

\begin{table}[ht!]
	\scriptsize
	\centering
	\caption{Parameter set}
	\label{tab:parameters}
	  \begin{tabular}{ cccccc }
	  
	  Parameter &$\alpha$ &$\beta$ &$s$ &$\sigma_{avg}$ &$tf$ \\\cline{1-6}\noalign{\vskip 1mm}
	  Value &$1$ &$.5$ &$0.02~m$ &$0.04~cm$ &$1$
	  
	  \end{tabular}	
\end{table}

\subsection{Performance Under Noisy Initial Guess}
\label{sec:noisy_init_guess}

This experiment is designed to show the robustness of the proposed approach under different noise conditions affecting the initial guess, and different directional scale discrepancies. For each \acrshort{ugv} point cloud, we estimate an accurate ground truth non-rigid transform by manually selecting the correct point-to-point correspondences  with the related \acrshort{uav} cloud. We generate random initial alignments between maps by manually adding noise, with different orders of magnitude, to the ground truth heading, translation, and scale. Then, we align the clouds with the sampled initial alignments by using (i) the proposed approach; (ii) a modified version of the proposed approach by moving from the \acrshort{exg} + \acrshort{dsm} environment representation to an RGB one (iii) a non-rigid standard \acrshort{icp}, (iv) the \acrlong{cpd} (\acrshort{cpd}) method \cite{myronenko2010}, (v) a state-of-the art \acrlong{goicp} (\acrshort{goicp}) \cite{yang2016}, and with standard sparse visual feature matching approaches \cite{Bay2008,10.1109/ICCV.2011.6126544,Calonder:2010:BBR:1888089.1888148}, applied as a data association front-end to our method in place of the proposed \acrshort{ldof} based data association (\secref{sec:ldof}): in the last cases, we exploit only the \acrshort{exg} channel of the grid maps (\secref{sec:multimodal_grid}). An alignment is considered valid if: $e_t <= 0.05~m$, $e_r <= 0.1~rad$, and $e_s <= 2.5\%$.

\begin{table*}[ht!]

	\centering
    \setlength{\tabcolsep}{2.9pt} 
    \scriptsize
	\caption{Registration accuracy comparison among the proposed approach, the non-rigid \acrshort{icp}, the \acrshort{cpd}~\cite{myronenko2010}, and the \acrshort{goicp}~\cite{yang2016} systems. The table reports, for each cell, the average accuracy among all the successful registrations with a specific initial anisotropic scaling error.}
	\label{tab:err_statistic}
	  \begin{tabular}{ ccccccccc }
	  \makecell{crop type} &approach &\makecell{registration err.\\ (trans/ros/scale) \\ scale error 0\%} &\makecell{registration err.\\ (trans/ros/scale) \\ scale error 5\%} &\makecell{registration err.\\ (trans/ros/scale) \\ scale error 10\%} &\makecell{registration err.\\ (trans/ros/scale) \\ scale error 15\%} &\makecell{registration err.\\ (trans/ros/scale) \\ scale error 20\%} &\makecell{registration err.\\ (trans/ros/scale) \\ scale error 25\%} &\makecell{registration err.\\ (trans/ros/scale) \\ scale error 30\%}\\\cline{1-9}\noalign{\vskip 1mm}

	  \multirow{7}{*}{\parbox[c]{1cm}{\centering Soybean}}

	  &\ourmethod &$0.03m/0.03^{\circ}/-$ &$0.03m/0.04^{\circ}/1.9\%$ &$0.04m/0.05^{\circ}/2.0\%$ &$0.04m/0.04^{\circ}/2.1\%$ &$0.03m/0.04^{\circ}/2.2\%$ &$0.05m/0.04^{\circ}/2.2\%$ &$0.05m/0.05^{\circ}/2.3\%$\\
	  &\acrshort{icp} &$0.03m/0.07^{\circ}/-$ &$0.05m/0.08^{\circ}/2.4\%$ &$0.04m/0.09^{\circ}/2.4\%$ &fail &fail &fail &fail\\
	  &\acrshort{cpd}~\cite{myronenko2010} &$0.02m/0.03^{\circ}/-$ &$0.04m/0.07^{\circ}/2.1\%$ &$0.03m/0.08^{\circ}/2.3\%$ &$0.03m/0.08^{\circ}/2.4\%$ &fail &fail &fail\\
	  &\acrshort{goicp}~\cite{yang2016} &$0.03m/0.06^{\circ}/-$ &- &- &- &- &- &-\\
	  &SURF~\cite{Bay2008} &$0.02m/0.04^{\circ}/-$ &$0.03m/0.04^{\circ}/2.2\%$ &$0.05m/0.06^{\circ}/2.4\%$ &fail &fail &fail &fail\\
	  &	~\cite{10.1109/ICCV.2011.6126544} &$0.02m/0.04^{\circ}/-$ &$0.04m/0.04^{\circ}/2.2\%$ &$0.05m/0.05^{\circ}/2.3\%$ &fail &fail &fail &fail\\
	  &FAST+BRIEF~\cite{Calonder:2010:BBR:1888089.1888148} &$0.04m/0.05^{\circ}/-$ &$0.05m/0.06^{\circ}/2.3\%$ &$0.05m/0.07^{\circ}/2.4\%$ &fail &fail &fail &fail\\
	  \cline{1-9}\noalign{\vskip 1mm}

	  \multirow{7}{*}{\parbox[c]{1cm}{\centering Sugar Beet 10m}}

	  &\ourmethod &$0.03m/0.04^{\circ}/-$ &$0.03m/0.04^{\circ}/2.1\%$ &$0.04m/0.04^{\circ}/2.0\%$ &$0.05m/0.06^{\circ}/2.0\%$ &$0.05m/0.07^{\circ}/2.3\%$ &$0.05m/0.1^{\circ}/2.3\%$ &$0.05m/0.1^{\circ}/2.4\%$\\
	  &\acrshort{icp} &$0.04m/0.05^{\circ}/-$ &$0.05m/0.07^{\circ}/2.1\%$ &$0.05m/0.09^{\circ}/2.4\%$ &fail &fail &fail &fail\\
	  &\acrshort{cpd}~\cite{myronenko2010} &$0.03m/0.04^{\circ}/-$ &$0.04m/0.05^{\circ}/2.1\%$ &$0.04m/0.06^{\circ}/2.2\%$ &$0.05m/0.09^{\circ}/2.4\%$ &fail &fail &fail\\
	  &\acrshort{goicp}~\cite{yang2016} &$0.02m/0.05^{\circ}/-$ &- &- &- &- &- &-\\
      &SURF~\cite{Bay2008} &$0.03m/0.04^{\circ}/-$ &$0.03m/0.04^{\circ}/2.1\%$ &$0.04m/0.07^{\circ}/2.3\%$ &fail &fail &fail &fail\\
	  &ORB~\cite{10.1109/ICCV.2011.6126544} &$0.02m/0.03^{\circ}/-$ &$0.03m/0.03^{\circ}/2.2\%$ &$0.05m/0.06^{\circ}/2.4\%$ &fail &fail &fail &fail\\
	  &FAST+BRIEF~\cite{Calonder:2010:BBR:1888089.1888148} &$0.02m/0.04^{\circ}/-$ &$0.02m/0.03^{\circ}/2.1\%$ &$0.05m/0.06^{\circ}/2.3\%$ &fail &fail &fail &fail\\
	  \cline{1-9}\noalign{\vskip 1mm}
	  
      \multirow{7}{*}{\parbox[c]{1cm}{\centering Sugar Beet 20m}}

	  &\ourmethod &$0.03m/0.03^{\circ}/-$ &$0.04m/0.03^{\circ}/2.0\%$ &$0.04m/0.04^{\circ}/2.2\%$ &$0.05m/0.05^{\circ}/2.1\%$ &$0.05m/0.08^{\circ}/2.2\%$ &$0.05m/0.09^{\circ}/2.4\%$  &$0.05m/0.1^{\circ}/2.4\%$\\
	  &\acrshort{icp} &$0.05m/0.06^{\circ}/-$ &$0.05m/0.09^{\circ}/2.3\%$ &fail &fail &fail &fail &fail\\
	  &\acrshort{cpd}~\cite{myronenko2010} &$0.04m/0.05^{\circ}/-$ &$0.05m/0.07^{\circ}/2.3\%$ &$0.05m/0.08^{\circ}/2.4\%$ &$0.05m/0.1^{\circ}/2.5\%$ &fail &fail &fail\\
	  &\acrshort{goicp}~\cite{yang2016} &$0.04m/0.05^{\circ}/-$ &- &- &- &- &- &-\\      
	  &SURF~\cite{Bay2008}   &$0.03m/0.04^{\circ}/-$ &$0.04m/0.05^{\circ}/2.1\%$ &$0.04m/0.06^{\circ}/2.4\%$ &fail &fail &fail &fail\\
	  &ORB~\cite{10.1109/ICCV.2011.6126544}    &$0.04m/0.05^{\circ}/-$ &$0.04m/0.05^{\circ}/2.2\%$ &$0.04m/0.05^{\circ}/2.4\%$ &fail &fail &fail &fail\\
	  &FAST+BRIEF~\cite{Calonder:2010:BBR:1888089.1888148} &$0.03m/0.04^{\circ}/-$ &$0.04m/0.05^{\circ}/2.1\%$ &$0.05m/0.07^{\circ}/2.4\%$ &fail &fail &fail &fail\\
	  \cline{1-9}\noalign{\vskip 1mm}
	  
	  \multirow{7}{*}{\parbox[c]{1cm}{\centering Winter Wheat}}

	  &\ourmethod &$0.04m/0.02^{\circ}/-$ &$0.04m/0.03^{\circ}/2.0\%$ &$0.04m/0.05^{\circ}/2.1\%$ &$0.04m/0.04^{\circ}/2.2\%$ &$0.05m/0.08^{\circ}/2.3\%$ &$0.05m/0.09^{\circ}/2.4\%$ &$0.05m/0.1^{\circ}/2.4\%$\\
	  &\acrshort{icp} &$0.04m/0.07^{\circ}/-$ &$0.04m/0.08^{\circ}/2.2\%$ &$0.05m/0.10^{\circ}/2.5\%$ &fail &fail &fail &fail\\
	  &\acrshort{cpd}~\cite{myronenko2010} &$0.04m/0.05^{\circ}/-$ &$0.04m/0.05^{\circ}/1.9\%$ &$0.04m/0.05^{\circ}/2.1\%$ &$0.05m/0.09^{\circ}/2.3\%$ &fail &fail &fail\\
	  &\acrshort{goicp}~\cite{yang2016} &$0.03m/0.07^{\circ}/-$ &- &- &- &- &- &-\\
      &SURF~\cite{Bay2008} &$0.03m/0.06^{\circ}/-$ &$0.03m/0.05^{\circ}/2.2\%$ &$0.04m/0.06^{\circ}/2.4\%$ &fail &fail &fail &fail\\
	  &ORB~\cite{10.1109/ICCV.2011.6126544} &$0.04m/0.05^{\circ}/-$ &$0.04m/0.04^{\circ}/2.1\%$ &$0.04m/0.06^{\circ}/2.3\%$ &fail &fail &fail &fail\\
	  &FAST+BRIEF~\cite{Calonder:2010:BBR:1888089.1888148} &$0.03m/0.07^{\circ}/-$ &$0.04m/0.06^{\circ}/2.3\%$ &$0.05m/0.05^{\circ}/2.4\%$ &fail &fail &fail &fail\\
	  \cline{1-9} 
	  \end{tabular}	
\end{table*}

The results are illustrated in Fig.~\ref{fig:correct_reg_rate}. The proposed approach significantly outperforms the other approaches, ensuring an almost $100\%$ success registration rate up to a scale error of $25\%$, and a high probability of succeeding even with a $30\%$ scale error. The ICP-based registration methods \cite{myronenko2010,yang2016}, due to the absence of structural 3D features on the fields, fall into local minima with high probability. The closest methods, in terms of robustness, are based on local feature matching \cite{Bay2008,10.1109/ICCV.2011.6126544,Calonder:2010:BBR:1888089.1888148}, succeeding in the registration procedure up to a scale error magnitude of $10\%$. While analyzing the results, however, we verified that, unlike our method, these methods provide a larger number of wrong, incoherent point associations, and such a problem is clearly highlighted for increasing scale deformations above 20\% and rotations above 0.1 radians. The superior robustness is also confirmed for noisy initial guesses: unlike the other methods, our approach guarantees a high successful registration rate for a translational error up to $5$ meters, and an initial heading error up to $11.5$ degrees, enabling it to deal with most errors coming from a GPS or \acrshort{ahrs} sensor. Our method generalizes well over the different datasets, showing the capability to deal with different crop species, crop growth stages (i.e., the winter wheat crop is in an advanced growth stage compared to the soybean and sugar beet), soil conditions, and point cloud resolution (from different \acrshort{uav} altitudes). An additional important outcome is the higher alignment probability obtained with the \acrshort{exg}/\acrshort{dsm} representation over the RGB one.

In Table~\ref{tab:inliers_percentage}, we report a comparison between the inliers percentages when using the visual (i.e., the \acrshort{exg} or the RGB) and the geometric terms in the cost function of \eqref{eq:cost_function}. Most of the information is carried by the visual term, especially by the \acrshort{exg}, while the sole geometric term is not able to provide valid results. Nevertheless, when combined, the latter acts  as a strong outliers rejection term, improving the robustness properties of the registration procedure. This is true especially for the sugar beet dataset, where the inliers percentage increases quite significantly.

\begin{table}[ht!]
	\scriptsize
	\centering
	\setlength{\tabcolsep}{3pt} 
	\caption{Inliers percentage comparison when changing data terms in the \acrshort{ldof} cost function.}
	\label{tab:inliers_percentage}
	  \begin{tabular}{ ccccc }
	  \multicolumn{1}{c}{} & \multicolumn{4}{c}{Descriptor Type (\% inliers)}\\\cline{2-5}\noalign{\vskip 1mm}
	  Crop Type    &RGB                  &\acrshort{exg}               &Depth           &\acrshort{exg} + Depth \\\cline{1-5}\noalign{\vskip 1mm}
	  Soybean      &$11.7\%\pm4.3\%$     &$53.2\%\pm14.9\%$ &$0.2\%\pm0.1\%$ &$\mathbf{54.5\%\pm13.2\%}$  \\
	  Sugar Beet    &$49.2\%\pm11.9\%$    &$64.1\%\pm12.8\%$ &$0.4\%\pm0.2\%$ &$\mathbf{68.1\%\pm13.6\%}$  \\
	  Winter Wheat &$22.9\%\pm9.7\%$     &$51.8\%\pm17.4\%$ &$0.1\%\pm0.1\%$ &$\mathbf{52.4\%\pm16.7\%}$  \\
	  \cline{1-5} 
	  \end{tabular}	
\end{table}

\subsection{Accuracy Evaluation}

To evaluate the accuracy of the proposed registration approach, we compare our results with the ground truth parameters and, by using all the successful registrations, we compute the average accuracy for each crop type and approach. The results are summarized in Tab.~\ref{tab:err_statistic}, and are sorted in increasing order of scale error.

\begin{figure}[ht!]
     \centering
     \includegraphics[width=.99\columnwidth, height=4.3cm]{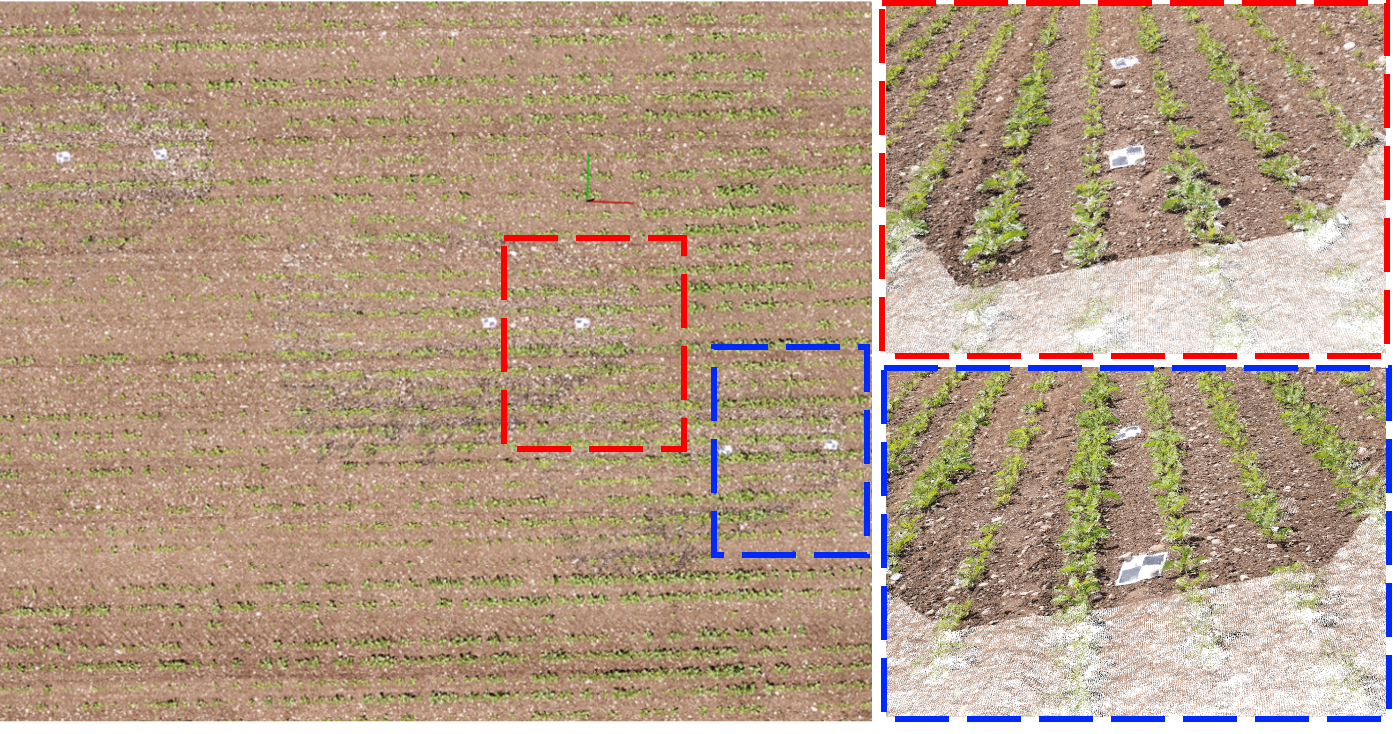}
     \caption{Qualitative registration results seen from aerial (left) and ground point-of-views. In the former, the UGV clouds are indistinguishable from the UAV, proving the correctness of the registration. Conversely, in the latter, the UGV clouds are clearly visible due to their higher points density.}
     \label{fig:qualit_view}
\end{figure}

On average, our method results in a lower registration error as compared to all the other evaluated methods for the same scale error. The difference in the registration error is even more pronounced when comparing the \textit{Sugar Beet 10m} against \textit{Sugar Beet 20m} datasets. Indeed, due to the higher sparseness of the points in the latter, all the other methods tend to perform slightly worse than they do with the \textit{Sugar Beet 10m}. Conversely, our method results in almost the same registration error magnitudes, showing that it correctly deals with the different densities of the initial colored point clouds. We also report some qualitative results in \figref{fig:qualit_view} 

\subsection{Runtime Evaluation}
\label{sec:comp_eff_eval}

We recorded the average, maximum, and minimum computational time for all tested methods over 100 successful registrations, reporting these values in Tab.~\ref{tab:timings_tab}. The method requiring the biggest computational effort is \acrshort{goicp}. The proposed approach requires half the computational time as compared to \acrshort{goicp}, but turns out to be quite slow compared to the custom-built \acrshort{icp}, and, in general, to all the other matching approaches. Fig.~\ref{fig:comp_time_pie} shows the runtime percentages for the proposed approach. The biggest component of the computational effort is required to extract the geometric features (i.e., the \acrshort{fpfh} features), meaning that the total computational time might be reduced by switching to a less time consuming 3D feature or by using only the visual term.

\begin{table}[ht]
\begin{minipage}[b]{0.5\columnwidth}
    \includegraphics[width=\columnwidth]{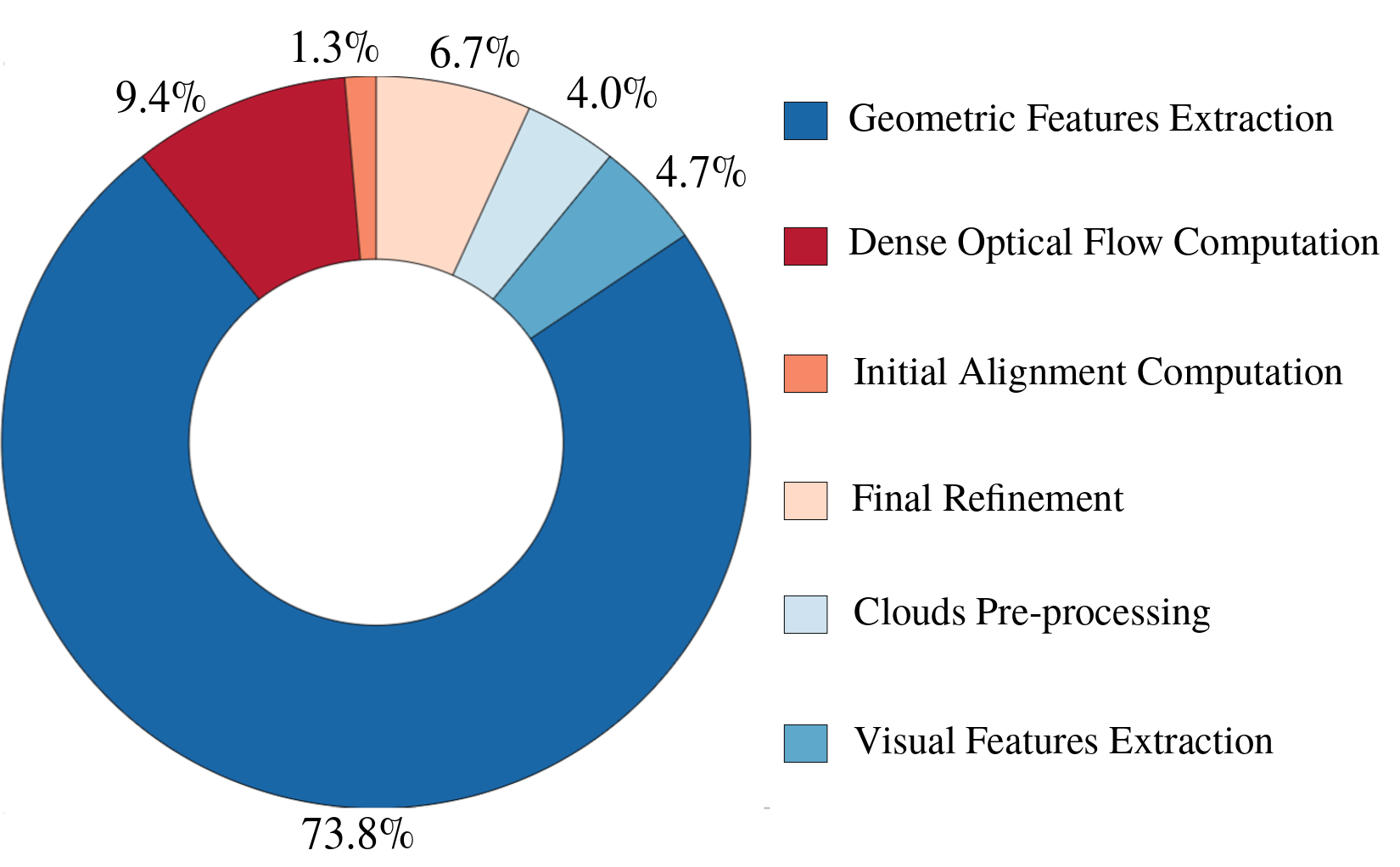}
    \captionof{figure}{Average percentage of the total runtime for different parts of the AgriColMap pipeline.}
    \label{fig:comp_time_pie}
\end{minipage}\hfill
\begin{minipage}[b]{0.45	\columnwidth}
	\scriptsize
	\centering
	\setlength{\tabcolsep}{1.5pt} 
    \begin{tabular}{ cccc }	  
        \multicolumn{1}{c}{} & \multicolumn{3}{c}{ \textbf{Runtime [sec]} }\\\cline{2-4}\noalign{\vskip 1mm}
        \multicolumn{1}{c}{} &Min &Max &Avg  \\\cline{1-4}\noalign{\vskip 1mm}
        $\ourmethod$          &63.7 &118.6 &79.8  \\\noalign{\vskip 1mm}
        \acrshort{icp}   &2.1  &10.6  &4.5   \\\noalign{\vskip 1mm}
        \acrshort{cpd}   &4.9  &23.2  &8.2   \\\noalign{\vskip 1mm}
        \acrshort{goicp} &5.3  &689.2 &193.1 \\\noalign{\vskip 1mm}
        SURF~\cite{Bay2008}             &4.6  &7.2   &5.3   \\\noalign{\vskip 1mm}
        ORB~\cite{10.1109/ICCV.2011.6126544}              &3.9  &6.7   &4.8   \\\noalign{\vskip 1mm}
        FAST+BRIEF~\cite{Calonder:2010:BBR:1888089.1888148}       &3.7  &6.4   &4.5   \\\cline{1-4}\\
    \end{tabular}	
    \caption{Runtime comparison.}
    \label{tab:timings_tab}
\end{minipage}\hfill
\end{table}

\section{CONCLUSIONS}

We addressed the cooperative \acrshort{uav}-\acrshort{ugv} environment reconstruction problem in agricultural scenarios by proposing an effective way to align heterogeneous 3D maps. Our approach is built upon a multimodal environment representation that uses the semantics and the geometry of the target field, and a data association strategy solved as a \acrshort{ldof} problem. We reported a comprehensive set of experiments, proving the superior robustness of our approach against standard methods. An open-source implementation of our system and the acquired datasets are made publicly available with this paper.

\section{ACKNOWLEDGEMENTS}
The authors would like to thank Hansueli Zellweger from the ETH Plant Research Station in Eschikon, Switzerland for preparing the fields, managing the plant life-cycle and treatments during the entire growing season. The authors would also like to thank Dr. Frank Liebisch from the Crop Science Group at ETH Z\"{u}rich for the helpful discussions.
\addtolength{\textheight}{1cm}
\balance

\bibliographystyle{IEEEtran}
\bibliography{ral_2018_agricolmap}

\end{document}